%% file: Main.tex
\documentclass[sigconf]{acmart}
\usepackage{multirow}
\usepackage{booktabs}
\usepackage{amsthm,amsmath}
\usepackage{mathrsfs}
\usepackage{graphicx}
\usepackage{color,xcolor}

\AtBeginDocument{%
  \providecommand\BibTeX{{%
    \normalfont B\kern-0.5em{\scshape i\kern-0.25em b}\kern-0.8em\TeX}}}

\acmConference[WWW '23]{ACM Web Conference 2023}{April 30--May 04, 2023}{Austin, TX, US}
\acmPrice{15.00}
\acmISBN{978-1-4503-XXXX-X/18/06}

\acmSubmissionID{1097}

\copyrightyear{2023}
\acmYear{2023}
\setcopyright{acmlicensed}\acmConference[WWW '23]{Proceedings of the ACM Web Conference 2023}{May 1--5, 2023}{Austin, TX, USA}
\acmBooktitle{Proceedings of the ACM Web Conference 2023 (WWW '23), May 1--5, 2023, Austin, TX, USA}
\acmPrice{15.00}
\acmDOI{10.1145/3543507.3583232}
\acmISBN{978-1-4503-9416-1/23/04}

\begin{document}

\title{CapEnrich: Enriching Caption Semantics for Web Images via Cross-modal Pre-trained Knowledge}


\author{Linli Yao}
\affiliation{%
  \institution{Renmin University of China}
  \city{Beijing}
 \country{China}
}
\email{linliyao@ruc.edu.cn}

\author{Weijing Chen}
\affiliation{%
  \institution{Renmin University of China}
  \city{Beijing}
 \country{China}
}
\email{chenweijing17@ruc.edu.cn}

\author{Qin Jin}
\authornote{Corresponding author.}
\affiliation{%
  \institution{Renmin University of China}
  \city{Beijing}
 \country{China}
}
\email{qjin@ruc.edu.cn}

\begin{abstract}
Automatically generating textual descriptions for massive unlabeled images on the web can greatly benefit realistic web applications, e.g. multimodal retrieval and recommendation.
However, existing models suffer from the problem of generating ``over-generic'' descriptions, such as their tendency to generate repetitive sentences with common concepts for different images.  These generic descriptions fail to provide sufficient textual semantics for ever-changing web images. Inspired by the recent success of Vision-Language Pre-training (VLP) models that learn diverse image-text concept alignment during pretraining,  we explore leveraging their cross-modal pre-trained knowledge to automatically enrich the textual semantics of image descriptions.  With no need for additional human annotations, we propose a plug-and-play framework, i.e CapEnrich,  to complement the generic image descriptions with more semantic details.  Specifically, we first propose an automatic data-building strategy to get desired training sentences, based on which we then adopt prompting strategies,  i.e. learnable and template prompts, to incentivize VLP models to generate more textual details. For learnable templates, we fix the whole VLP model and only tune the prompt vectors, which leads to two advantages: 1) the pre-training knowledge of VLP models can be reserved as much as possible to describe diverse visual concepts; 2) only lightweight trainable parameters are required, so it is friendly to low data resources. 
Extensive experiments show that our method significantly improves the descriptiveness and diversity of generated sentences for web images. 
The code is available at \href{https://github.com/yaolinli/CapEnrich}{https://github.com/yaolinli/CapEnrich}.
\end{abstract}




\settopmatter{printfolios=true}

\maketitle

\input{Introduction}

\input{Related_Work}

\input{Model}

\input{Experiment}

\input{Conclusion}

\noindent \textbf{ACKNOWLEDGMENTS}
This work was partially supported by National Natural Science Foundation of China (No. 62072462), National Key R\&D Program of China (No. 2020AAA0108600).
\bibliographystyle{ACM-Reference-Format}
\bibliography{Reference}

\appendix
\input{Appendix}

\end{document}

%% file: Introduction.tex
\section{Introduction}
\begin{figure}[t]
    \centering
    \includegraphics[width=0.9\linewidth]{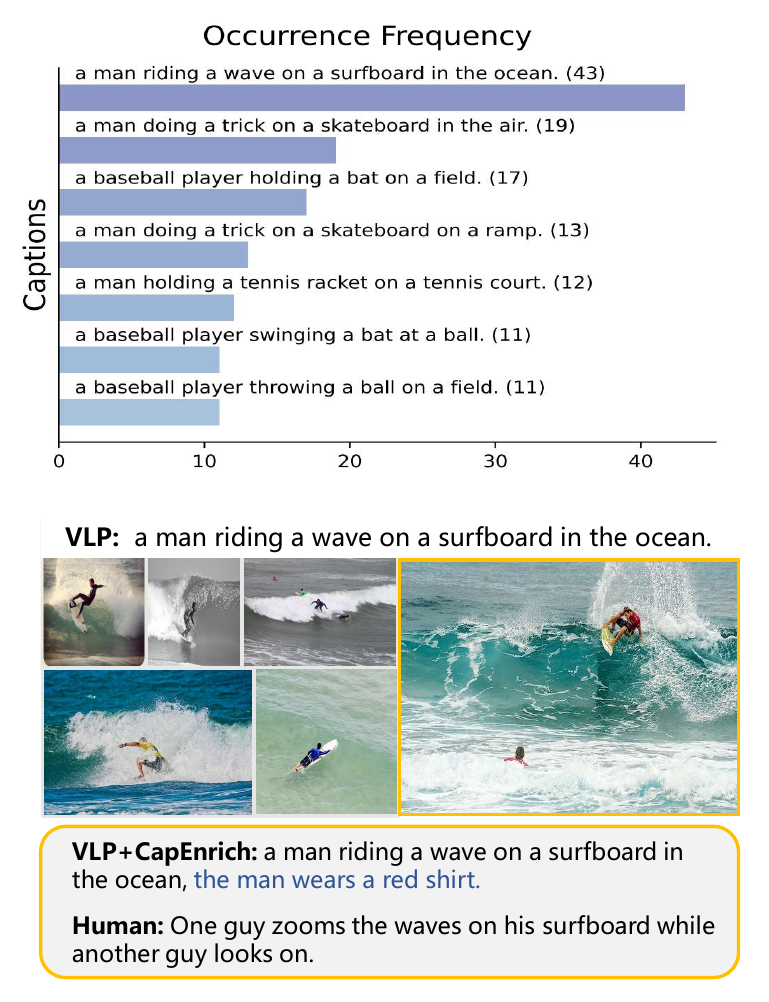}
    \vspace{-8pt}

    \caption{Occurrence frequency of some repetitive captions generated by a VLP model VinVL~\cite{vinvl} on MSCOCO test set.  A specific case of enriching the generated description of VinVL with more salient details via CapEnrich.}
    \label{fig:task_intro}
\end{figure}

Massive amount of unlabeled images are uploaded to the web every day from advertisements, online chatting and personal photo albums etc. Indexing and labeling them with textual descriptions can greatly benefit realistic applications, such as Content-based Image Retrieval and Recommendation~\cite{Hossain2019ACS}. 
Image captioning~\citep{Vinyals2015Show,Xu2015Show,rennie2017self}, which aims to automatically generate natural language descriptions given an unlabeled image, is exactly one technique suitable for this purpose. 
Recently, the quality of automatically generated image captions has been significantly boosted by the success of  Vision-Language Pre-training (VLP) models~\citep{uniter, clip, vinvl}. These VLP models are pre-trained with Internet-scale image-text pairs broadly crawled from the web, e.g. SBU Captions~\citep{sbu} and Conceptual Captions~\citep{cc3m, cc12m}. Based on the cross-modal pre-training knowledge, they achieve conspicuous performances on  downstream tasks with fine-tuning.

However, the modern image captioning models, even the VLP-based models, still suffer from the problem of generating ``overly-generic''~\cite{wang2020compare,wang2021group} captions, for example, the models tend to produce generic descriptions with common concepts for different images. As shown in Figure~\ref{fig:task_intro}, we observe that the state-of-the-art VLP-based model, VinVL~\citep{vinvl}, generates exactly the same caption ``\textit{a man riding a wave on a surfboard in the ocean}'' for 43 different images, and ``\textit{a man doing a trick on a skateboard in the air}'' for 19 different images in the MSCOCO~\cite{mscoco} test set. Due to the ``overly-generic'' problem, the existing image captioning models fail to provide detailed semantics for web images thus are insufficient to support realistic applications with diverse user demand, for example, fine-grained web image retrieval or recommendation. 

A straightforward solution to overcome the ``overly-generic'' problem is to obtain lots of desired rich human annotations and urge the existing models to learn from such data, which is highly costly and time-consuming. Some endeavors have been made to automatically generate more descriptive image captions with detailed semantics, such as designing additional rewards~\citep{liu2018show, luo2018discriminability, liu2019generating} or  integrating natural language inference (NLI) relations~\citep{NliCap}. However, their optimization schemes are designed upon a limited scope of training samples and neglect the advantages of the recent VLP models. Instead, our insight is that since VLP models learn diverse visual-textual concept alignments from Internet-scale  pre-training corpus, they should have the potential capacity to describe more visual details in addition to general contents. The key question is how we stimulate them to describe more details. 

In this paper, we shed light on a new perspective to leverage the cross-modal pre-training knowledge of VLP models to automatically enrich the generic image descriptions with more detailed semantics. We first conduct preliminary experiments to verify the intrinsic capacity  of VLP models to generate more detailed image captions. Based on the observation, we propose  an \textit{annotation-free}, \textit{light-weight} and \textit{plug-and-play} framework to leverage this capacity via prompting. As Figure~\ref{fig:task_intro}  shows, our goal is to automatically complement details (e.g. ``the man wears a red shirt'') for the generic caption (e.g. ``a man riding a wave on a surfboard in the ocean'') by leveraging the cross-modal pre-training knowledge. 
 
To achieve the \textit{annotation-free} goal, we first propose an automatic data-building strategy that aggregates multiple captions to form a more detailed one. We observe that in the caption benchmark datasets, each image already has multiple captions from different annotators with varied visual focus and narrative granularity. Therefore, we extract the various textual details (relations and attributes) from multiple sentences and aggregate them to obtain the desired data samples in the  format of  ``generic caption, details''.

Based on above limited new-format data, we adopt prompting strategies to stimulate VLP models to automatically generate more significant details in a \textit{light-weight} manner. To be specific, we append learnable prompt vectors to the end of the generic caption, and the training goal is to predict the detail words based on the visual semantics, the generic caption and the prompt vectors. During training, we fix the whole VLP model and only update the parameters of prompts for two advantages: 1) the pre-trained knowledge of VLP models can be reserved as much as possible to describe diverse visual concepts; 2) only lightweight trainable parameters are updated, which is friendly to low data resources. Our novel framework, consisting of the automatic data-building strategy and the prompting approaches, can be  \textit{plug-and-played} into any VLP backbones, enriching the caption decoding. Specifically, we demonstrate its effectiveness on two recent state-of-the-art VLP models, VinVL~\citep{vinvl} and X-VLM~\citep{xvlm}. 
In summary, the main contributions of this paper are:
\begin{itemize}
    \item To the best of our knowledge, this is the first attempt to automatically stimulate VLP models to enrich the generic captions with more significant semantics for web images.
    \item We propose a novel plug-and-play framework with an automatic data-building strategy and prompt-tuning approaches. It is annotation-free, light-weight and can be applicable to any VLP models.
    \item   Extensive experiments show that our framework gain remarkable improvements on the \textit{descriptive} metrics against state-of-the-art models, even comparable to human annotations.
    Meanwhile, we gain the ability of controllable generation by training class-specific prompts.
\end{itemize}

%% file: Related_Work.tex
\section{Related Work}

\noindent \textbf{Descriptive Image Captioning.}
Image captioning~\citep{Vinyals2015Show,rennie2017self,Xu2015Show} aims to describe image contents with natural language.
Previous works~\citep{liu2018show,liu2019generating, wang2021group} have pointed out that image captioning models trained  with cross-entropy loss and reinforcement learning rewards~\cite{rennie2017self} will  generate     ``overly-generic'' captions. 
Therefore, various works~\cite{dai2017contrastive, wang2021group} are proposed to improve the descriptiveness or distinctiveness of the captions. For descriptiveness, \citet{stack-cap} proposes a coarse-to-fine strategy and \citet{liu2019generating} utilizing visual paraphrases in a two-stage decoder. Recently, \citet{NliCap} leverages natural language inference relations among different annotations to obtain more descriptive captions.
For discriminability, contrastive learning~\citep{dai2017contrastive,luo2018discriminability,chen2018groupcap} and self-retrieve strategy~\citep{liu2018show, vered2019joint} are widely adopted to push the generated caption closer to the target image while pull it further away from  other images. Besides, \citet{wang2020compare} highlights the distinctive annotations by a weighted loss and \citet{wang2021group} emphasizes the unique objects of each image. 

However, all these works are built on the traditional captioning architecture, leading to an explicit upper bound that existing models are unlikely to generate finer grained sentences  than the annotated ones.  
In contrast, we take full advantage of the transferable pre-training knowledge of the latest vision-language pre-training models~\citep{ vinvl, xvlm}  to generate more detailed captions.


\noindent \textbf{Vision-Language Pre-training (VLP).}
Pre-training on large-scale image-text pairs and fine-tuning on downstream vision-language tasks has been a successful paradigm~\citep{uniter, clip, vinvl, xvlm}. 
\textcolor{black}{Based on the multi-layer transformer, pre-trained models learn to align diverse visual and textual concepts through self-supervised tasks. On the visual side, they normally take multiple object features~\cite{vinvl,oscar,uniter,Lu2019ViLBERTPT,Tan2019LXMERTLC,Wang2020VDBERTAU,Li2019UnicoderVLAU,Zhou2019UnifiedVP}, grid features~\cite{soho} or patches~\cite{Gao2020FashionBERTTA, Wang2021SimVLMSV, Kim2021ViLTVT} as input to acquire comprehensive visual semantics.}
For image captioning, pre-trained models are commonly adapted with a two-stage fine-tuning to generate accurate descriptions: first trained with Cross-Entropy (CE) loss and then optimized by reinforcement learning with CIDEr rewards~\citep{scst}. However, the descriptiveness of captions including significant visual details is neglected. In this paper, we aim to alleviate the ``overly-general'' caption problem and encourage VLP-based methods to enrich the original generic captions with more salient details.

\noindent \textbf{Prompt Learning.} NLP tasks have witnessed a prevalence of prompt engineering~\citep{radford2019language, brown2020language, raffel2020exploring, liu2021pre}. Different from fine-tuning that adapts pre-trained language models to downstream tasks, prompt learning adjusts the  downstream tasks to approach pre-trained models, thus can better leverage the pre-trained knowledge. The basic idea is designing instructional templates manually~\citep{brown2020language, schick2020exploiting} or automatically~\citep{li2021prefix,prompt-tuning} for pre-trained models to obtain desired output answer. It has the advantages of being parameter-efficient and performs brilliantly in zero-shot or few-shot scenarios. Latest works~\citep{tsimpoukelli2021multimodal, zhou2021learning,yao2021cpt} extend prompt learning to vision-language tasks and achieve remarkable results. The most related works are~\citep{cho2021unifying} and \citep{jin2021good}, which also probe prompt learning in downstream captioning tasks. However, they only apply hand-crafted prompts to refine the fine-tuning stage and improve the accuracy metrics. Instead, our goal is to stimulate VLP models to complement more non-trivial details using both hand-crafted and learnable prompts.

%% file: Model.tex

\section{Background}
In this section, we aim to answer whether VLP models can be directly stimulated to generate more semantic details beyond generic description. 
We first review the adaptation of VLP models to image captioning tasks via the fine-tuning paradigm.
We then examine the potential of VLP models to generate more details when stimulated by designed prompts. 

\subsection{\textcolor{black}{Adapting VLP models to image captioning }}
\label{sec:vlp_gen}
Building a VLP-based image captioning model typically consists of a three-phase process: 1) firstly pre-training the VLP model on Internet-scale image-text pairs with self-supervised tasks to align the cross-modal semantics, 
2) then fine-tuning the model on the image caption benchmark datasets to acquire generation ability with a cross-entropy (CE)
loss, 
and 3) finally optimizing the model with a reinforcement (RL) reward~\cite{scst} to improve the captioning metrics .
For presentation simplicity, we use the state-of-the-art VLP model VinVL~\cite{vinvl} as our default backbone unless otherwise specified. 
VinVL has a one-stream  architecture with a multi-layer cross-modal Transformer.
It takes image region features $V$ and word embedding sequences $W$ as input~\footnote{We omit the input object tags in VinVL model for presentation convenience.}. In the pre-training stage, it is optimized by a masked language modeling (MLM) task and a cross-modal contrastive learning task utilizing 8.85 million web-associated image-text pairs. In the fine-tuning stage, it adjusts the masked language modeling with a uni-direction attention mask to acquire the image caption generation ability. It masks out 15\% word tokens \{$w_{m}$\} in the text sequence with [MASK] tokens and predicts  the corresponding  token ids using output hidden states. The uni-directional MLM task is firstly optimized by a cross-entropy loss $\mathcal{L}_{\mathrm{CE}}$ and then a reinforcement learning loss $\mathcal{L}_{\mathrm{RL}}$~ with CIDEr rewards to further improve the captioning metrics.
\begin{equation}
\mathcal{L}_{\mathrm{CE}}=-\mathbb{E}_{ (V,W) \sim \mathcal{Z}} \left[ \log p\left(w_{m} \mid w_{\backslash m}, V \right) \right]
\end{equation}

\begin{equation}
\mathcal{L}_{\mathrm{RL}}=-\mathbb{E}_{ \hat{W} \sim \log p\left(W \mid V \right)} \left[ Reward(\hat{W}, V) \right]
\end{equation}

During inference, VinVL decodes a sentence auto-regressively by feeding [MASK] tokens and conditioning on visual contents. Since the generated sentences only describe the generic visual contents of an image as Figure~\ref{fig:task_intro} illustrated, we call these sentences directly decoded by VLP models as ``generic caption''.

\begin{figure}[t]
    \centering
    \includegraphics[width=\linewidth]{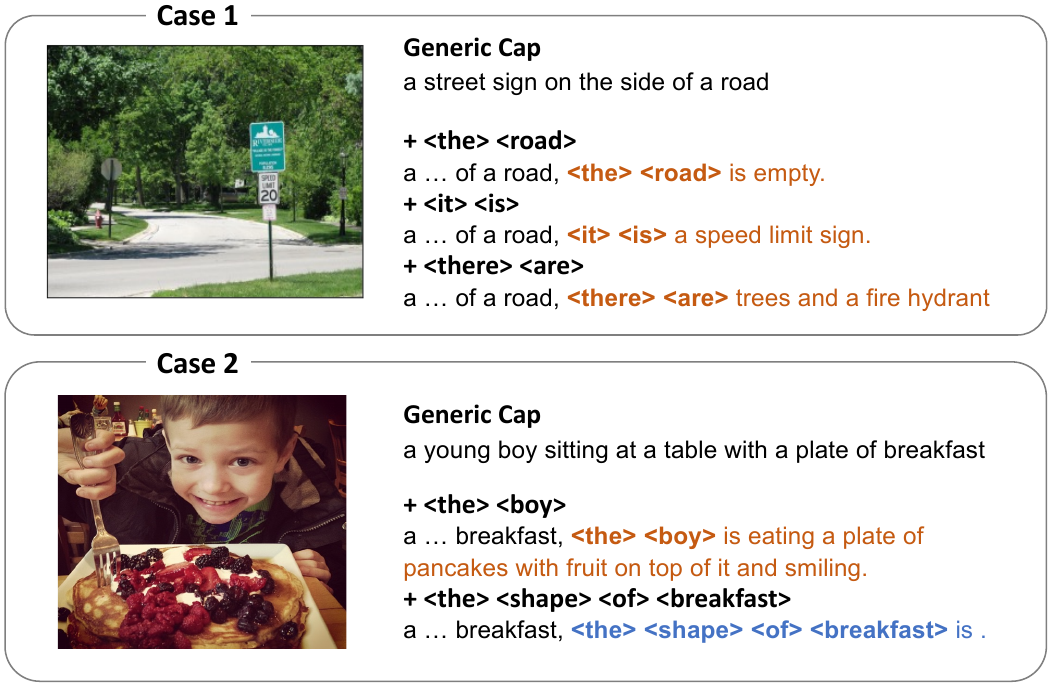}
    \vspace{-8pt}
    \caption{Cases of ``Template Prompts'' that guiding VinVL~\citep{vinvl} model to directly generate more detailed textual contents.
    }
    \label{fig:template_case}
\end{figure}

\begin{figure*}[t]
    \centering
    \includegraphics[width=0.99\linewidth]{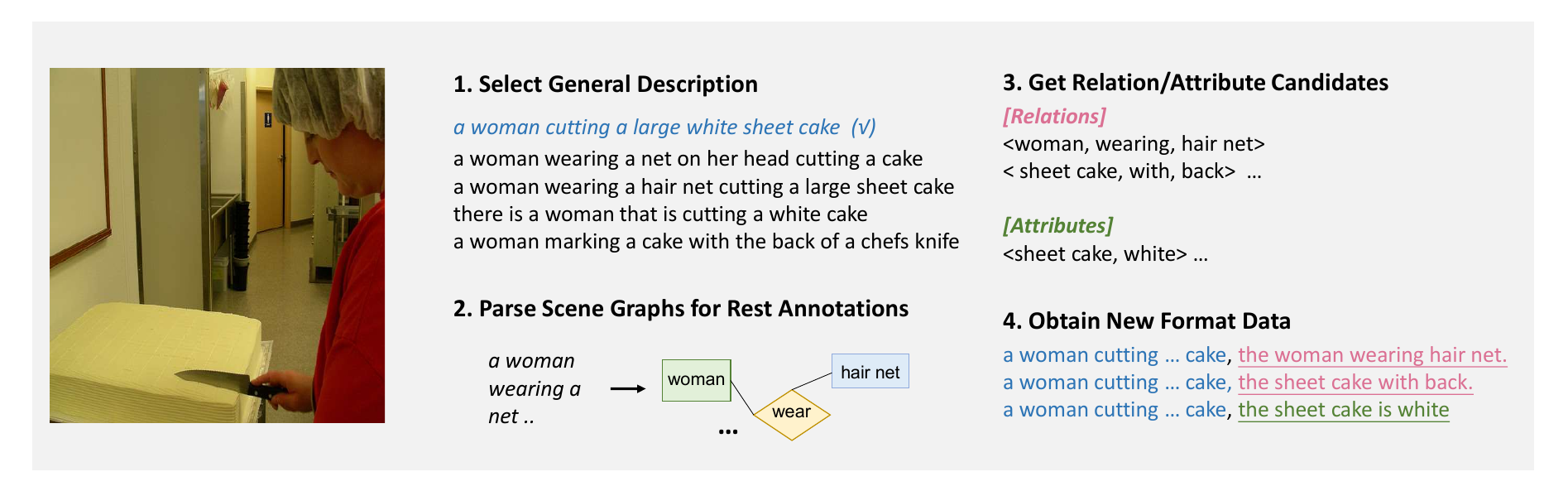}
    \vspace{-8pt}
    \caption{Illustration of the automatic data building strategy. We aggregate multiple annotated sentences and produce a more descriptive one in the format of ``generic caption, details'' following four steps.
    }
    \label{fig:data_construct}
\end{figure*}

\subsection{Can we stimulate VLP-based models to generate more detailed  caption?}
Based on the knowledge hierarchy, our human can naturally complement more details beyond the generic visual content for an unseen image. Inspired by the human cognitive capacity,  we therefore design textual templates as a guidance to find out whether VLP models can be triggered to generate more textual details directly.

Specifically, we design several contextual words as template prompts and append them at the end of the generic caption to encourage VLP models to generate more visual details. Take the ``Template Prompts'' case 1 in Figure~\ref{fig:template_case} as example, we locate the noun phrases \textit{``a road''} in the generic caption \textit{``a street sign on the side of a road''}. Then we use \textit{``$\langle\textrm{the}\rangle$ $\langle\textrm{road}\rangle$''} as the prompting words to form the template prompt as 
\textit{``a street sign on the side of a road, $\langle\textrm{the}\rangle$ $\langle\textrm{road}\rangle$''}. 
It is fed to the VLP models along with the image information to trigger the model to continuously decode new words. Serving as the history textual semantics, the template prompt stimulates VLP models to generate more details about ``the road''.

We empirically design different templates and analyze the newly decoded ``detail words'' by VinVL. More diverse hand-crafted templates details are described in the Appendix. 
We have following observations:
\begin{itemize}
    \item VinVL has the human-like capability to naturally complement more visual details for an unseen image with appropriate guidance. Cases are shown in Figure~\ref{fig:template_case} and quantitative results are presented in Experiments  (Section~\ref{sec:handcrafted}).

    \item VinVL can leverage the cross-modal pre-training knowledge, i.e. diverse visual-textual concept alignment, to describe more details with novel expressions not appeared in the downstream benchmark datasets. For example, in Figure~\ref{fig:template_case} case 2, VinVL complements the details ``pancakes with fruit on top of it'' for the ``breakfast''.
    
    \item The template prompts should be well-designed according to the prior knowledge and human labor in order to play an effective guiding role. Case 2 presents a failing case where the prompt ``<the> <shape> <of>'' can not stimulate more semantic details.
\end{itemize}

Based on these observations, we see that VLP-based models have decent potentials to supplement images with more semantic details, and the key lies in how to effectively stimulate the model. Therefore, we would like to propose an universal learnable framework that automatically encourages VLP models to generate more detailed descriptions, while leveraging the cross-modal pre-training knowledge as much as possible.

\section{Proposed Framework}

\begin{figure*}[t]
    \centering
    \includegraphics[width=0.9\linewidth]{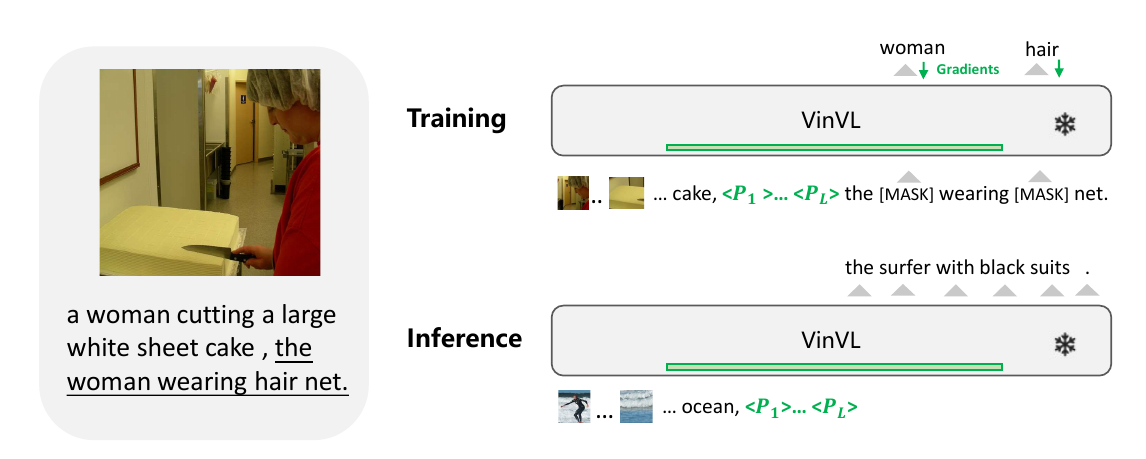}
    \vspace{-8px}
    \caption{Overview of the prompting strategies based on the VinVL~\citep{vinvl} model.  Given an image $\mathcal{I}$ and a generic description $\mathcal{G}$, our goal is to automatically enrich $\mathcal{G}$ with more meaningful details $\mathcal{D}$ (underlined). We append learnable prompt tokens $\{\left<P_1\right> \ldots \left<P_L\right>\}$ at the end of $\mathcal{G}$ as special textual tokens. The training objective is to predict the masked words in $\mathcal{D}$ based on the visual and history textual semantics, which endow the VinVL model with detail generation ability. During training, we fix the VinVL model and only update the learnable prompts for parameter-efficiency and reserving the pre-trained knowledge as much as possible.
    }
    \label{fig:model}
\end{figure*}

It is not trivial to incentivize VLP models to generate a descriptive caption including significant details at once without corresponding supervised annotations.  Referring to our human cognitive habits when describing visual content, that is, we  tend to first roughly describe the overall image and then add more details according to individual visual interests. We therefore propose a novel  framework (\textbf{CapEnrich}) for VLP models to automatically complement more significant details over their generic descriptions.
We first introduce an automatic data-building strategy (Section~\ref{sec:auto_data}) to obtain new-format data depicted in Figure~\ref{fig:data_construct}. Then we design prompting strategies (Section~\ref{sec:prompt}) to invoke the rich pre-trained knowledge in VLP models through lightweight training as illustrated in Figure~\ref{fig:model}.

\subsection{Problem Formulation}
Given an image $\mathcal{I}$, a generic description $\mathcal{G}$ can be naturally obtained by a VLP model after fine-tuning on the downstream image captioning task. Based on the visual information  $\mathcal{I}$ and the obtained generic caption  $\mathcal{G}$, our goal is to stimulate the VLP models to generate more significant textual details  $\mathcal{D}$. Finally, we aggregate both the generic caption and the additional textual details via post-processing, to obtain the paragraph-type descriptive captions denoted as $\mathcal{Y}$.
\begin{align}
\mathcal{G}&={W^{\mathcal{G}}}=\{w^\mathcal{G}_{1},\dots, w^\mathcal{G}_{N}\}, \nonumber \\
\mathcal{D}&={W^\mathcal{D}}=\{w^\mathcal{D}_{1},\dots, w^\mathcal{D}_{M}\}, \nonumber \\
\mathcal{Y}&=\{\mathcal{G}, \mathcal{D}\} = \{w^\mathcal{G}_{1},\dots, w^\mathcal{G}_{N}, w^\mathcal{D}_{1},\dots, w^\mathcal{D}_{M}\}
\end{align}

\subsection{Automatic Data Building}
\label{sec:auto_data}
To enable supervision for training the learnable prompts to invoke the VLP model to generate more details, we design an automatic procedure to build the required data in the format $\mathcal{Y}=\{\mathcal{G}, \mathcal{D}\}$. We observe that existing image captioning benchmark datasets have multiple textual annotations from different annotators for each image. Although annotators describe the overall content of the same image, they have distinct visual focus and narrative granularity. Therefore, we consider that it is possible to condense general caption and aggregate more details from these different annotations. This way we can build the supervision data for training the model to complement details without requiring additional human annotations.

Figure~\ref{fig:data_construct} depicts our proposed automatic data-building procedure and illustrates a specific case.
For an image $\mathcal{I}$ with $\textrm{n (n$\geq$2)}$ annotated captions $\mathcal{C}=\{c_{i}\}_{i\in[1,n]}$, we first find the shortest sentence from $\mathcal{C}$ as the generic  caption $\mathcal{G}$. Then we parse\footnote{https://github.com/vacancy/SceneGraphParser} the rest captions and extract the scene graphs $\{\mathcal{S}\}$. We can obtain  relation or attribute candidates from the scene graphs. If a relation or attribute candidate does not appear in $\mathcal{G}$, we consider it as a meaningful additional detail beyond generic content $\mathcal{G}$. We aggregate the selected caption candidates into a sub-sentence $\mathcal{D}$ using simple rules, and consequently form the new description data as $\{\mathcal{G},\mathcal{D}\}$.

\subsection{Detail Generation with Prompting}
\label{sec:prompt}
Given the image $\mathcal{I}$ and a generic description $\mathcal{G}$, we apply prompting strategies to stimulate VLP models to generate more semantic details $\mathcal{D}$. 
Considering that the state-of-the-art VLP model has mastered general captioning ability from pre-training and fine-tuning, we employ its raw inference output for the given image $\mathcal{I}$  as the generic description  $\mathcal{G}$ introduced in Section~\ref{sec:vlp_gen}.
We explore  learnable (soft) prompting to design a \textit{light-weight} and \textit{plug-and-play} framework to obtain $\mathcal{D}$ as illustrated in Figure~\ref{fig:model}.

Inspired by the Prompt Tuning work~\cite{prompt-tuning}, we adopt learnable prompts with the Fixed-VLP training strategy for two motivations: preserve the pre-training knowledge as much as possible and reduce the trainable parameters as much as possible.
Specifically, we append a series of task-specific continuous vectors $\mathcal{P}=\{\left<P_1\right> \ldots \left<P_L\right>\}$ to the end of the generic caption $\mathcal{G}$. They can be seen as additional special tokens, whose embeddings will be updated during training.
The training objective is the same as in the fine-tuning stage with CE loss, that is, masking and predicting the words in textual details $\mathcal{D}$ conditioned on $\mathcal{I}, \mathcal{G}, \mathcal{P}$. Note that our goal is to endow VLP models with the detail generation capability, therefore, we only mask the words in $\mathcal{D}$ rather than in $\mathcal{G}$.
\begin{equation}
\mathcal{L}_{\mathrm{PT}}=-\mathbb{E}_{V,W \sim \mathcal{Z}} \log p\left(w^{\mathcal{D}}_{m} \mid w^{\mathcal{D}}_{\backslash m}, W^{\mathcal{P}}, W^{\mathcal{G}}, V \right)
\end{equation}

The continuous prompt vectors can be initialized randomly or through high-frequency word embeddings. During inference, we can decode the detail words auto-regressively by feeding the  generic description, the trained prompt vectors and a starting [MASK] token.

The learnable prompts play an indicative  role in  the decoding process of VLP models and serve as the prior textual semantics to encourage the VLP models to decode more detail-related words. Since any VLP-based image captioning model has autoregressive generation capability, our overall framework  can be plug-and-played into any architecture.

\subsection{Aggregation in Post-processing}
We conduct post-processing in the final aggregation step to insure the quality of the enriched description $\mathcal{Y}=\{\mathcal{G},\mathcal{D}\}$. For each target image, our framework can generate multiple descriptive captions via different template or learnable prompts. We  filter meaningless  captions, e.g. the semantics of enriched details are repeated in the generic caption, with following steps. Firstly we filter out structural-incomplete sentences by Spacy. Then we calculate the semantic similarity $sim(.)$ between descriptions and the target image using CLIP.  We reserve those descriptive captions whose $sim(\mathcal{G}\mathcal{D}, \mathcal{I}) > sim(\mathcal{G}, \mathcal{I}) $, which means the details part has effective additional textual semantics.
For metric evaluation, we select the most descriptive caption.

%% file: Experiment.tex
\begin{table*}[tbp]
\begin{center}
\caption{ Comparison with the state-of-the-arts on MSCOCO dataset from two aspects: \textit{descriptiveness} and \textit{accuracy}. C-S and RefC-S refer to CLIPScore and RefCLIPScore respectively. We adopt the proposed plug-and-play framework to two state-of-the-arts VLP models, VinVL and X-VLM. LP denotes learnable prompts. X-VLM has two
settings including exploiting 4M pre-training corpus and
16M pre-training corpus.
}
\vspace{-8pt}
\begin{tabular}{c|l|l@{ }l@{ }ll@{ }l@{ }l|l@{ }l@{ }l}
\toprule
\multirow{3}{*}{} &
  \multirow{3}{*}{\textbf{Method}} &
  \multicolumn{6}{c|}{\textbf{\textit{Descriptiveness}}} &
  \multicolumn{3}{c}{\textbf{\textit{Accuracy}}} \\ 
 &
   &
  \multicolumn{3}{l}{\textbf{CLIP} \textrm{\textcolor{black}{(Hard Retrieval Pool)}}} &
  \multicolumn{3}{l|}{\textbf{VSE++} \textrm{\textcolor{black}{(Naive Retrieval Pool)}}} &
  \multicolumn{1}{l@{}}{\multirow{2}{*}{\textbf{C-S}}} &
  \multicolumn{1}{@{}l@{}}{\multirow{2}{*}{\textbf{RefC-S}}} &
  \multicolumn{1}{@{}l}{\multirow{2}{*}{\textbf{SPICE}}} \\ 
 &
   &
  \textbf{R@1} &
  \textbf{R@5} &
  \textbf{R@10} &
  \textbf{R@1} &
  \textbf{R@5} &
  \textbf{R@10} &
  \multicolumn{1}{l@{}}{} &
  \multicolumn{1}{l@{}}{} &
  \multicolumn{1}{l}{} \\ 
  \toprule
  
1& DiscCap~\citeyearpar{luo2018discriminability} &
  - &
  - & 
  - &
  24.4 &
  54.9 &
  70.1 &
  74.1 &
  80.5 &
  21.2 
   \\
2& StackCap~\citeyearpar{stack-cap} &
  - &
  - & 
  - &
  21.9 &
  49.7 &
  63.7 &
  - &
  - &
  20.9 
   \\
3& CBtwCap~\citeyearpar{wang2020compare} &
  - &
  - & 
  - &
  26.5 &
  58.0 &
  71.3 &
  - &
  - &
  23.0 
   \\
4& NliCap~\citeyearpar{NliCap} &
2.9   &
9.3   &
15.6   &
29.1   &
60.5   &
73.9   &
75.5   &
81.6   &
22.9   \\
5& GdisCap~\citeyearpar{wang2021group} &

  3.5 &
  10.8 &
  16.9 &
  31.1 &
  62.7 &
  76.0 &
  75.8 &
  81.7 &
  23.0 \\

  \midrule
6& $\textrm{VinVL}$~\citeyearpar{vinvl} &

  5.8 &
  15.6 &
  23.2 & 
  36.4 &
  67.7 &
  80.5 &
  77.4 &
  83.1 &
  25.1 \\
7& + $\textrm{LP}$ & 
  \textbf{\textcolor{blue}{9.3} \textcolor{gray}{($+{3.5}$)}} &
  \textbf{\textcolor{blue}{22.6} \textcolor{gray}{($+{7.0}$)}} &
  \textbf{\textcolor{blue}{31.3} \textcolor{gray}{($+{8.1}$)}} &
  \textbf{\textcolor{blue}{38.0} \textcolor{gray}{($+{1.6}$)}} &
  \textbf{\textcolor{blue}{70.5} \textcolor{gray}{($+{2.8}$)}} &
  \textbf{\textcolor{blue}{82.5} \textcolor{gray}{($+{2.0}$)}} &
  \textbf{\textcolor{blue}{79.2} \textcolor{gray}{($+{1.8}$)}} &
    83.3 \textcolor{gray}{($+{0.2}$)} &
    25.1 \\
   \midrule
8& $\textrm{X-VLM}_{4m}$~\citeyearpar{xvlm} & 
  4.9 &
  13.0 &
  19.9 &
  31.0 &
  61.8 &
  74.6 &
  76.8 &
  82.8 &
  24.2  \\
9& + $\textrm{LP}$ & 
  8.8 \textcolor{gray}{($+{3.9}$)} &
  21.6 \textcolor{gray}{($+{8.6}$)} &
  31.1 \textcolor{gray}{($+{11.2}$)} &
  37.6 \textcolor{gray}{($+{6.6}$)} &
  69.6 \textcolor{gray}{($+{7.8}$)} &
  81.8 \textcolor{gray}{($+{7.2}$)} &
  79.2 \textcolor{gray}{($+{2.4}$)} &
  \textbf{\textcolor{blue}{83.9} \textcolor{gray}{($+{1.1}$)}} &
  25.2 \textcolor{gray}{($+{1.0}$)}  \\

10& $\textrm{X-VLM}_{16m}$~\citeyearpar{xvlm}& 
  4.4 &
  13.1 &
  19.2 &
  28.8 &
  59.4 &
  73.4 &
  76.5 &
  82.5 &
  23.9  \\
11& + $\textrm{LP}$ & 
  8.1 \textcolor{gray}{($+{3.7}$)} &
  20.4 \textcolor{gray}{($+{7.3}$)} &
  29.4 \textcolor{gray}{($+{10.2}$)} &
  37.0 \textcolor{gray}{($+{8.2}$)} &
  69.2 \textcolor{gray}{($+{9.8}$)} &
  81.5 \textcolor{gray}{($+{8.1}$)} &
  78.9 \textcolor{gray}{($+{2.4}$)} &
  83.8 \textcolor{gray}{($+{1.3}$)} &
  \textbf{\textcolor{blue}{25.3} \textcolor{gray}{($+{1.4}$)}}  \\
  \midrule
12& \textcolor{black}{Human} &
  7.9  & 
  19.9 & 
  27.5 &
  30.3 &
  59.4 &
  72.4 &
  77.6 &
  81.9 &
  20.9 \\
 
  \bottomrule
\end{tabular}
\label{tab:sota}
\end{center}

\end{table*}

\section{Experiments}

\subsection{Experiment Settings}
\textbf{\textcolor{black}Datasets.} 
We adopt the automatic-data building strategy on two widely-used image captioning benchmark datasets, MSCOCO~\cite{mscoco} and Flickr30K~\cite{flickr30k}. Each image in the two datasets is crawed from Flickr  and annotated with five captions.
Following previous works, we adopt the widely used Karpathy~\cite{Karpathy2017DeepVA} split setting. MSCOCO dataset contains around 120,000 images, of which 5,000 images are used for validation, 5,000 images for testing and the rest for training. Flickr30K dataset contains around 31,000 images and we use 1000 images for validation and testing respectively. Based on the two datasets, We construct 356,824 and 103,734 new-format training samples respectively via our automatic data-building strategy.

\textbf{\textcolor{black}Evaluation Metrics.} 
We evaluate the caption quality from two main aspects: \textit{accuracy} and \textit{descriptivesness}. Traditional  metrics such as BLEU~\cite{bleu} and CIDEr~\cite{cider} are N-gram based algorithms, which are sensitive to the whole sentence structure~\cite{liu2019generating} (further discussion can be found in Appendix~\ref{sec:metric}). Therefore, for the \textit{accuracy} aspect, we report the SPICE~\cite{spice}, CLIPScore and RefCLIPScore~\cite{clipscore} to mitigate the structural impacts. CLIPScore is a reference-free metric that directly calculates the semantic similarity between image and text. In addition, RefCLIPScore also emphasizes the semantic similarities between reference annotations and the generated caption.
To evaluate the \textit{descriptiveness}, we adopt a self-retrieval strategy following previous works~\cite{liu2019generating,NliCap} based on the observation that more descriptive captions with significant details often lead to more precise self retrieval, which refers to retrieving the target image from a set of similar images given the generated caption. We report the refined R@K score using CLIP~\cite{clip} as the retrieval model on a well-designed more challenging candidates set (denoted as Hard Retrieval Pool), which is elaborated in Appendix. Besides, we also report the R@K scores on the commonly used  retrieval protocol (denoted as Naive Retrieval Pool) with  VSE++~\cite{vse++} for a fair comparison.To further analyse the diversity of generated image descriptions we choose widely used Div-n\cite{div-n}, mBLEU\cite{div-n} and Self-CIDEr\cite{selfcider} as diversity metrics.

\textbf{\textcolor{black}Implementation Details.} 
If not specified, we use $\textrm{VinVL}_{base}$ as the VLP backbone for image captioning.  For prompt tuning, we train the learnable vectors with learning rate 3e-4 and batch size 48. For full model training baseline (introduced in Section~\ref{sec:finetuning-baseline}), we set learning rate as 3e-6 and batch size as 48. We initialize the learnable prompts via two ways: 1) random initialization from a zero-mean Gaussian distribution, or 2) initialization from specified word embeddings. We compare the prompt in various length such as [1,2,4,8] and set to 2 as default. The maximum tuning epoch is set to 30 and the beam size for inference is 5. To emphasize the descriptiveness of generated captions, we select the best checkpoint by CLIP retrieval score R@1 on the validation set.

Since our framework can generate multiple enriched descriptions for an image, we select the most distinctive one, which has the highest similarity score with the target image, to compute the evaluation metrics. When ensembling multiple descriptions guided by different learned prompts, we also utilize the CLIP retrieval score to choose a best sentence from N candidates.

\subsection{Full-Model Training Baseline}
\label{sec:finetuning-baseline}
A straightforward way to endow VinVL with the capability of generating more details is to finetune it on the above automatically built data samples $\{\mathcal{G},\mathcal{D}\}$. 
Given the visual features $V$ and generic description embedding $W^{\mathcal{G}}$, it aims to predict the masked words in $\mathcal{D}$. With a uni-direction attention mask, the $W^{\mathcal{G}}$ serves as the history textual semantics. 
\begin{equation}
\mathcal{L}_{\mathrm{FT}}=-\mathbb{E}_{ (V,W) \sim \mathcal{Z}} \left[\log p\left(w^{\mathcal{D}}_{m} \mid w^{\mathcal{D}}_{\backslash m}, W^\mathcal{G}, V \right)\right]
\end{equation}
During fine-tuning, we apply a low learning rate to update the whole parameters $\phi$ of VinVL. We take such full-model training method as our baseline to compare with the enhanced model with prompting strategy.

\subsection{Comparison with State-of-the-Arts}

Table \ref{tab:sota} compares our model with other state-of-the-art descriptive captioning works, including DiscCap~\cite{luo2018discriminability}, StackCap~\cite{stack-cap}, CBtwCap~\cite{wang2020compare}, GdisCap~\cite{wang2021group} and NliCap~\cite{NliCap}. These models adopt the encoder-decoder captioning architecture without pre-training and aim to improve the descriptiveness of captions. We also report the  performance of human annotations as the upper bound for reference.

The VLP models, VinVL~\cite{vinvl} and X-VLM~\cite{xvlm}, have demonstrated superior performance than other models on both \textit{accuracy} and \textit{descriptiveness} metrics (row 6\&8 vs rows in the first block). However, in terms of descriptiveness, there is still a significant gap with the groundtruth (VinVL 5.8 vs Human 7.9 on CLIP R@1). Our prompting framework standing on the shoulder of the giant (the VLP models), further improves the descriptiveness of generated captions and archives new state-of-the-art performance (9.3 on CLIP R@1), which even surpasses the average groundtruth (7.9 on CLIP R@1). Remarkable improvements across the different VLP architectures, i.e. VinVL and X-VLM, further demonstrate the superiority and flexibility of our framework.
The accuracy metrics of SPICE and RefCLIPScore, which measure the consistency between the generated caption and each reference annotation, are also enhanced.  Since not every reference is descriptive enough, the improvements on SPICE and RefC-S are relatively smaller. In conclusion, our model achieves the best R@K, SPICE, CLIPScore and RefCLIP-S scores, proving its ability to generate more accurate and descriptive captions.

\begin{figure}[tbp]
    \centering
    \includegraphics[width=0.9\linewidth]{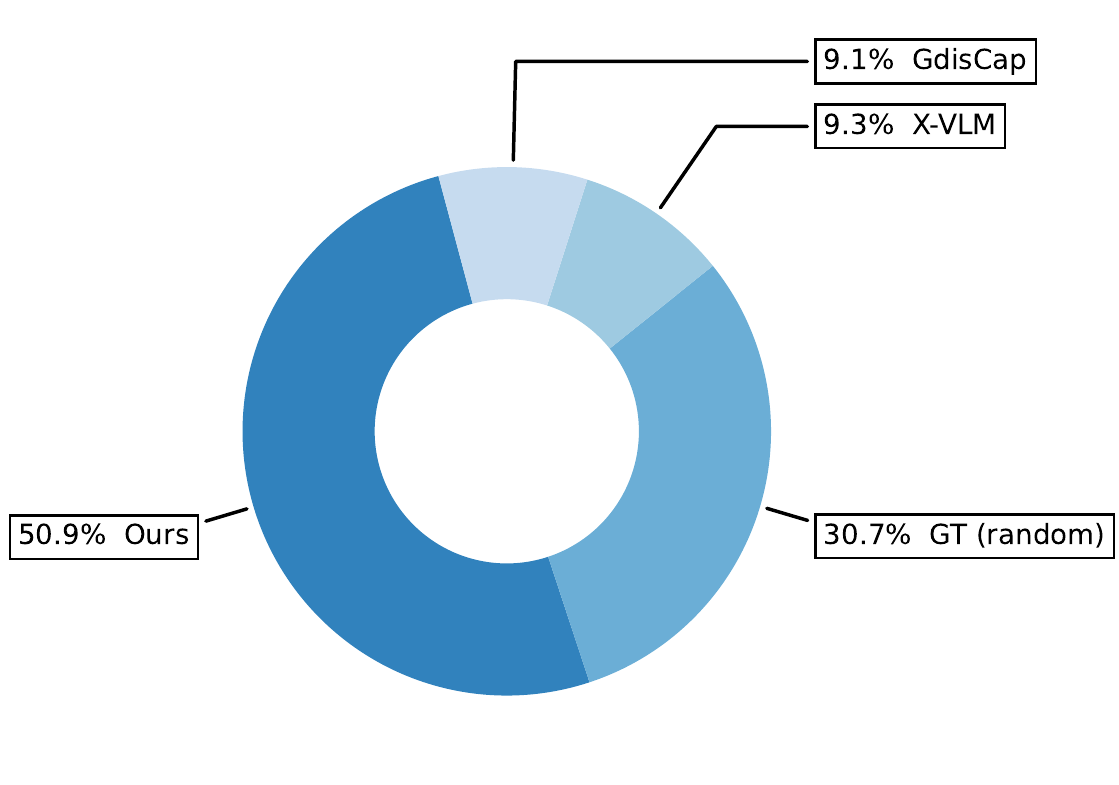}
    \vspace{-8pt}
    \caption{User study results. Visualization the frequency of being selected as the most descriptive caption.}
    \label{fig:user_study_pie}
\end{figure}

\noindent \textbf{User Study.}
Due to the limitation of automatic metrics, we further conduct a human evaluation as shown in Figure~\ref{fig:user_study_pie}. We recruit 18 participants to evaluate 50 randomly sampled cases from MSCOCO. For each case, participants are asked to select the accurate and the most descriptive caption among four candidates (more details are presented in Appendix).  Our model is the most favored one, significantly outperforming the annotation randomly selected from the groundtruth, which again proves the effectiveness of our model.

\begin{figure}[t]
    \centering
    \includegraphics[width =  \linewidth]{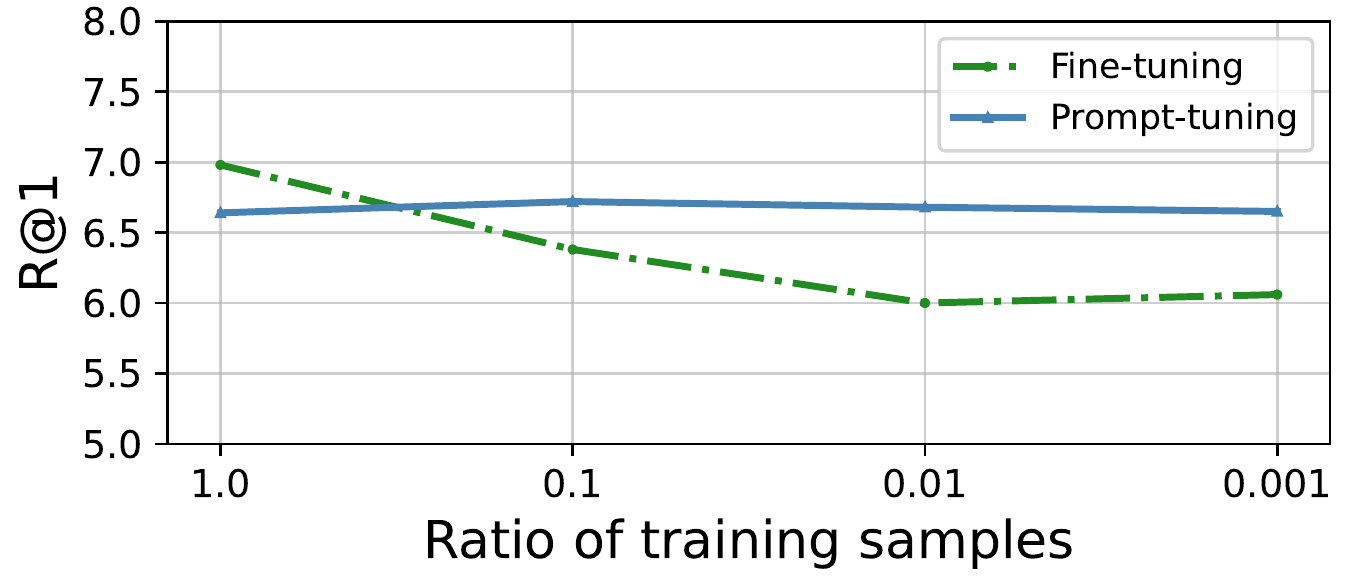}
    \caption{ Performance of full-model fine-tuning and prompt-tuning on MSCOCO with reduced training sizes.}
    \label{fig:data_ratio}
\end{figure}

\begin{table}[t]
\begin{center}
\caption{Performance comparison on CLIP R@K between the full-model training baseline (FT) and learnable prompting (LP) on MSCOCO. \#P is short for number of prompts.}
\vspace{-8pt}
\begin{tabular}{l|l|c|rrr}
\toprule
& \textbf{Tuning Method} & \textbf{\#P} & \textbf{R@1}  & \textbf{R@5}  & \textbf{R@10} \\ \midrule

1& VinVL &- &5.8 &15.6 &23.2   \\

2 & \multirow{1}{*}{VinVL w/ FT (88M)}  & - & 7.0 & 18.4 & 26.7 \\
3 & VinVL + $\textrm{LP}$  (1.5K)  & 1               & 6.8 & 17.8 & 25.8 \\
4 & VinVL + $\textrm{LP}$  (12K)   & 8 & 9.3 & 22.6 & 31.3 \\
\midrule
5 & Human & - &7.9 &19.9 &27.5 \\

\bottomrule
\end{tabular}
\label{tab:soft_results}
\end{center}
\end{table}

\subsection{Learnable Prompting Performance}
Table~\ref{tab:soft_results} compares the learnable prompting performance with the full-model training (finetuning) baseline.
The single learnable prompting achieves comparable performance as fine-tuning (6.8 vs 7.0 on R@1) although it only requires less than 0.01\% trainable parameters (1.5K vs 88M). 
When training N (N=8) groups of learnable prompts separately with different hyper-parameters and ensembling them during inference, it can significantly boost the performance (9.3 vs 6.8 on R@1) with a slight increase in parameters, even outperforming the average groundtruth performance (7.9 on R@1).
Moreover, Figure~\ref{fig:data_ratio} shows the advantage of prompting strategy with low data resources, especially when utilize only 0.001\% data sample.

\subsection{Zero-shot Detail Generation Performance}
\label{sec:handcrafted}
We further quantify the zero-shot detail generation performance of VinVL model. As shown in Table~\ref{tab:hd_results}, appending hand-crafted template prompts with post-processing boosts the descriptiveness of captions (row 2 vs row 1). It demonstrates that the VLP model can naturally continue to generate more semantic details if stimulated with extra context word phrases (specific designs are presented in Appendix~\ref{sec:appendix_template}). 
Notably, the performance with  template prompts (11.3 on R@1) is superior to the average level of groundtruth captions (7.9 on R@1, row 5 in Table~\ref{tab:soft_results}) without any additional training or parameters. This further demonstrates that prompting can effectively leverage the transferable pre-trained knowledge in VLP models to downstream captioning tasks. 
Comparing row 2 and row 4 in Table~\ref{tab:hd_results}, we get another interesting  observation, that is, directly fine-tuning on the small-corpus new data will hurt the inherent descriptive caption generation ability of VinVL (drops from 11.3 to 9.1 on R@1). In contrast, our prompting strategies  can maintain the pre-trained knowledge of VLP models as much as possible by freezing the whole parameters.

\begin{table}[tbp]
\begin{center}
\caption{CLIP retrieval performance of template prompts on MSCOCO dataset. }
\vspace{-8pt}
\begin{tabular}{l|l|rrr}
\toprule
 & \textbf{Model} &  \textbf{R@1}   & \textbf{R@5}  & \textbf{R@10}  \\ \midrule
1 & VinVL                     & 5.8  & 15.6 & 23.2  \\ 
2 & + $\textrm{template}$     & 11.3 & 26.2 & 36.2 \\
\midrule
3 & VinVL w/ FT & 7.0  & 18.4 & 26.7   \\ 
4 & + $\textrm{template}$     &9.1 &23.0 &32.1 \\ 
\bottomrule
\end{tabular} 
\label{tab:hd_results}
\end{center}
\end{table}

\begin{table}[t]
\begin{center}
\caption{Performance comparison on Flickr30K dataset.}
\vspace{-8pt}
\begin{tabular}{l|cccc}
\toprule
\textbf{Method}         & \textbf{R@1}   & \textbf{R@5}   & \textbf{R@10} & \textbf{SPICE}  \\ \midrule
VinVL  & 23.3 & 45.4 & 57.1 & 21.7      \\
+ $\textrm{LP}$  & \textbf{29.2} & \textbf{54.9} & \textbf{67.2} &  \textbf{22.3}    \\ 
\bottomrule
\end{tabular}
\label{tab:flickr30k}
\end{center}
\end{table}

\begin{table}[]
\begin{center}
\caption{Diversity Performance on MSCOCO dataset with five generated captions.}
\vspace{-8pt}
\begin{tabular}{l|cccc}
\toprule
  \textbf{Method}       & \textbf{Div-1}$\uparrow$ & \textbf{Div-2}$\uparrow$ & \textbf{Self-CIDEr}$\uparrow$ & \textbf{mBLEU4}$\downarrow$ \\ \midrule
VinVL   & 25.3 & 33.8 & 59.7 & 80.8      \\
w/ CapEnrich   & \textbf{25.8} & \textbf{36.9} & \textbf{62.4} & \textbf{73.3}       \\
\bottomrule
\end{tabular}
\label{tab:diversity}
\end{center}
\end{table}

\begin{figure*}[!htbp]
    \centering
    \includegraphics[width =   0.9\linewidth]{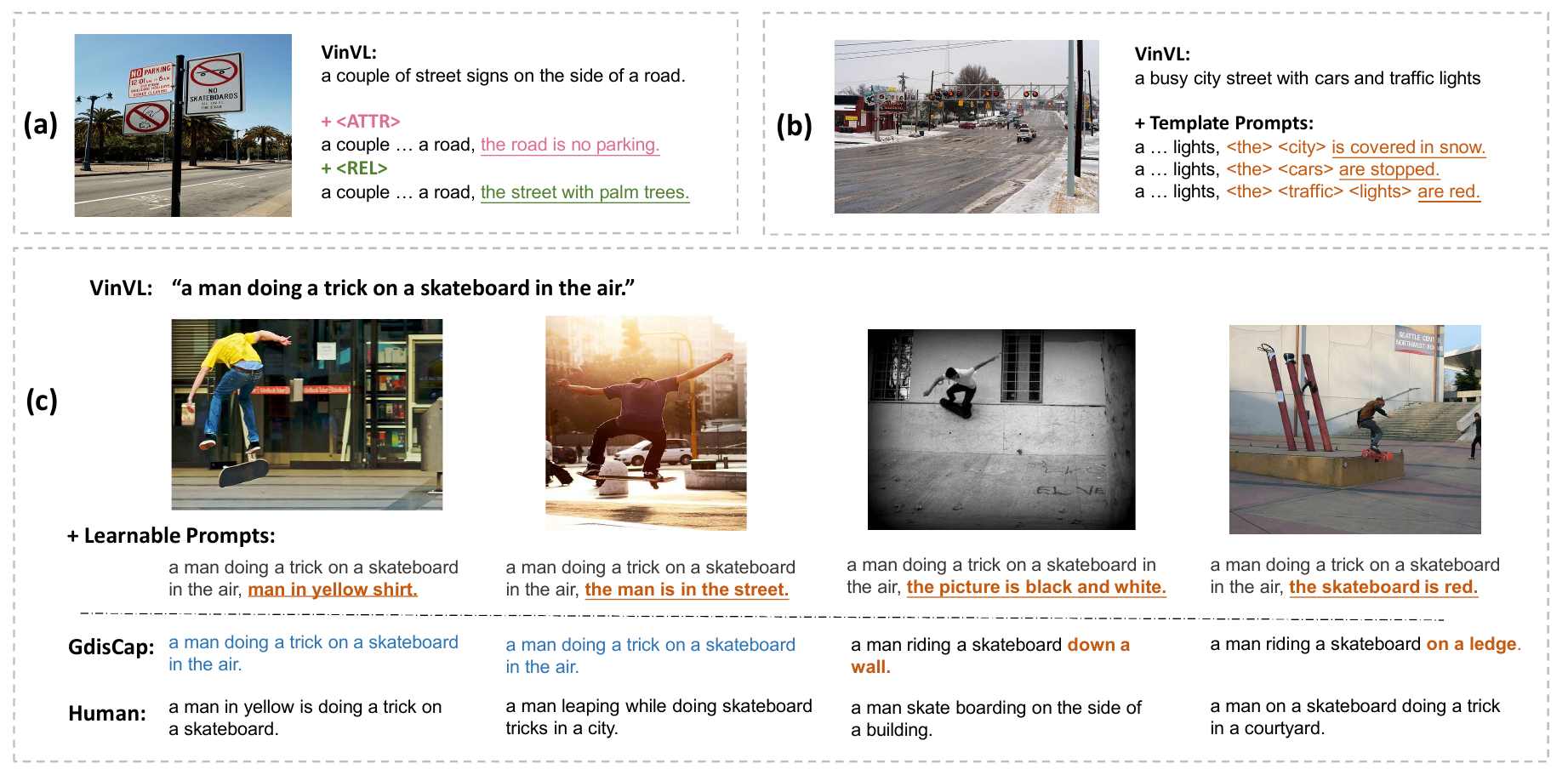}
    \vspace{-7pt}
    \caption{ Visualization of  captioning cases on MSCOCO test set. The newly generated details are \underline{underlined}. Our framework can enrich the generic caption generated by VinVL with more salient details.}
    \label{fig:cases}
\end{figure*}


\section{Further Analysis}
\textbf{Results on Different Datasets.}
To verify the extendability of our framework, we also report results on Flickr30K dataset in Table~\ref{tab:flickr30k}. Appending learnable prompts further boosts the \textit{descriptiveness} metric R@1 from 23.3 to 29.2 and the \textit{accuracy} metric SPICE from 21.7 to 22.3. It demonstrates that our plug-and-play framework is applicable to various benchmark datasets.

\noindent \textbf{Diversity of Generated Descriptions.}
We report the diversity metrics in the Table~\ref{tab:diversity}. Our CapEnrich framework can prompt a VLP model to generate multiple descriptive sentences with different templates or learned vectors. The diversity of captions can be significantly improved.

\noindent \textbf{Controllable Detail Generation.}
One side benefit of our proposed framework is that if we train class-specific learnable prompts, we can realize controllable detail generation.
To be specific, we split the automatically constructed data into two types: \textit{Attribute} and \textit{Relation}, according to scene graph labels. Then we train two groups of prompt vectors ($\langle\textrm{ATTR}\rangle$ and $\langle\textrm{REL}\rangle$) using the corresponding data respectively. During inference, we can utilize the  prompts to control the generated details, as illustrated in Figure~\ref{fig:cases} (a).

\noindent \textbf{Qualitative Studies.}
Figure~\ref{fig:cases} visualizes some generated caption examples. Case (b) represents details complemented  via template prompting ``the city''.   
In case (c), the original VinVL model generates the same caption about ``skateboarding'' for four different images. While appending learnable prompts can enrich more salient and significant details including the appearance of the man and the skateboard, the location and the overall visual effect.

\vspace{-2px}

%% file: Conclusion.tex
\vspace{-5px}
\section{Discussion and Conclusion}
In our CapEnrich framework, we position learnable prompts for newly enriched details at the end of the sentences  rather than in the middle, which leads to unconventional textual structures. There are two main reasons: 1) imitating the cognitive habits of humans in a ``first generic then detailed'' manner; 2) fitting the  auto-regressive decoding process which can obtain the entire contexts of generic captions and prompts served as previous tokens.
We argue that the semantic enrichment is the first concern in the web applications and our experiments show that modern image retrieval models like CLIP can handle the various sentence structure.

In summary, we  propose a brave new perspective that encourages VLP models to generate more descriptive captions for web images with salient and significant details, in order to better support the web applications, e.g. fine-grained image indexing and retrieval. Specifically, we introduce a novel  framework including an automatic-data building strategy and a prompt-based training method that is \textit{annotation-free}, \textit{light-weight}, and can be \textit{plug-and-played} to any VLP models and any benchmark datasets.
Extensive experiments on different captioning datasets and with different VLP backbones demonstrate the effectiveness and flexibility of our framework. 
\textcolor{black}{In the future, we will extend the flexible CapEnrich framework to controllable detailed caption generation,  to satisfy the different visual interests and caption purposes of users. Our prompting framework is feasible to integrate external user intent
contexts into the learnable prompt vectors at an instance level.}

%% file: Appendix.tex
\clearpage

\section{Appendix}

\begin{table}[thbp]
\begin{center}
\begin{tabular}{@{}l|l|r|rrr}
\toprule
Dataset                    & Retrieval Pool & \#Imgs & R@1  & R@5  & R@10 \\ \midrule
\multirow{2}{*}{MSCOCO}    & Naive          & 5,000    & 27.4 & 53.6 & 65.6  \\  
                           & Hard           & 34,259   & 5.8  & 15.6 & 23.2 \\ \midrule
\multirow{2}{*}{Flickr30K} & Naive          & 1,000    & 55.4 & 83.8 & 90.8 \\ 
                           & Hard           & 7,413    & 23.3 & 45.4 & 57.1  \\ \bottomrule
\end{tabular}
\end{center}
\caption{Image-to-text retrieval performance comparison on CLIP R@K between the naive retrieval pool and hard retrieval pool using the generated captions from original VinVL model as queries.}
\label{tab:clip_retrieval}
\end{table}

\subsection{Over-generic Problem in VLP Models}
\textcolor{black}{
The ``over-generic'' problem is an ingrained problem of modern captioning models due to the Maximum Likelihood Estimation(MLE) training objective. They tend to generate common words/concepts that occur frequently in ground truth sentences in order to achieve a higher overall likelihood score. Instead, producing rare/unique words that describe diverse visual regions will result in lower scores.  VLP models will also get stuck in the ingrained ``over-generic'' problem when fine-tuning on the captioning datasets with the MLE loss. Therefore, their pre-training knowledge of associating diverse visual-textual concepts will also be confined when generating a sentence at once. 
 Inspired by the human cognitive habits of ``first generic then detailed'', we simplify the problem 
 ``generating a detailed caption at once'' to a new manner ``first producing generic contents then enriching more details''. Conditioned on the generic contents and suitable prompts, VLP models can naturally describe more diverse visual details of the target image.}

\subsection{CLIP-based Self-Retrieval Metric}
\label{sec:retrieve}
We improve the widely-used self-retrieval scores, i.e. R@K, from two aspects: the retrieval model and the retrieval pool. On the one hand, we replace the  retrieval model VSE++ with the state-of-the-art image-text retrieval model CLIP to improve the retrieval precision.  On the other hand, we propose a more challenging candidates set with more similar images. As shown in Table~\ref{tab:clip_retrieval},  CLIP has achieved high text-to-image R@1 score  on MSCOCO test set and Flickr30K test set using the generated captions by original VinVL model. Therefore, a more challenging retrieval set with more similar images is necessary to reflect the relative descriptiveness improvement of generated captions. 

\begin{figure*}[t]
    \centering
    \includegraphics[width=0.9\linewidth]{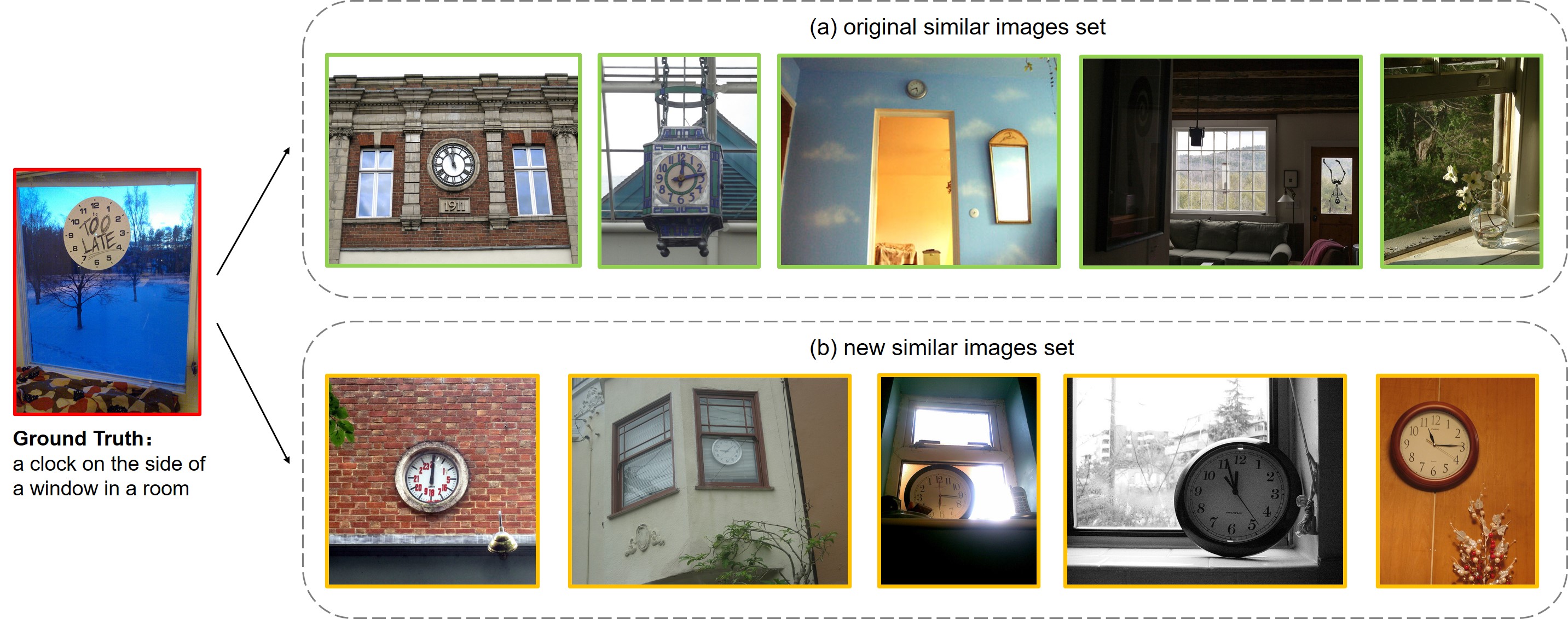}
    \caption{A case of original similar images set and new similar images set}
    \label{fig:similar images set}
\end{figure*}

We construct a new challenging images candidate set (denoted as hard retrieval pool) through the following processing stages. 1) We introduce additional large-scale unlabeled images ${\mathcal{R}}$ including the 2015 test images and 2017 unlabeled images from MSCOCO. We ensure that the extra images have no overlap with the existing captioning benchmark datasets. 2) For each target image in the benchmark test set, we retrieve top K similar images from ${\mathcal{R}}$ according to text-image and image-image similarity. 3) Finally we aggregate the overall retrieved similar images and obtain a new candidate set with larger size and more challenging candidates. 
Table~\ref{tab:clip_retrieval} shows the comparison of R@K results on the naive retrieval pool and the hard retrieval pool.
Figure~\ref{fig:similar images set} shows a case comparing similar images from the original candidate set and  the new set. The new set is clearly more challenging than the original set. 

\subsection{\textcolor{black}{Diverse Template Prompt Design}}
\label{sec:appendix_template}
We design different template prompts to explore the zero-shot detail generation capability of VLP models. As shown in Table~\ref{tab:handcraft}, the \textbf{base} version of template prompts is ``the + noun'' and the  \textbf{diversity} version considers more aspects  (Attribute, Number, Orientation, Weather and Others) according to the human prior observed from the MSCOCO dataset. Table~\ref{tab:hd_results_all} compares the performance of different templates.
 From row 2 to row 3 (or row 5 to row 6), a remarkable performance improvement (from 7.1 to 11.3 on R@1) indicates that well-designed prompts can better excavate the detail generation potential of VinVL.

\begin{table}[htbp]
\begin{center}
\begin{tabular}{l@{\qquad}l}
\toprule
\textbf{Type}         & \textbf{ Templates } \\ \midrule
Base                                   & \textrm{the X}                 \\ \hline \hline
\multirow{2}{*}{\textit{Attribute}}   & \textrm{the man/woman wears}        \\ 
                                       & \textrm{the color of X is}     \\ \hline
\textit{Number}       & \textrm{the number of X is }   \\ \hline
\textit{Orientation}                  & \textrm{on the right/left/top of X} \\ \hline
\textit{Weather } & \textrm{the weather is }        \\ \hline
\multirow{2}{*}{\textit{Others}} & \textrm{it is }                     \\  
                                       & \textrm{there is/are  }    \\ \bottomrule
\end{tabular}
\caption{Designed prompt templates including \textbf{base} version and \textbf{diverse} version (\textit{Attribute, Number, Orientation, New Objects} and \textit{Environment}). X is the placeholder of appeared nouns from the generic caption, e.g. ``table''.}
\label{tab:handcraft}
\end{center}
\end{table}

\begin{table}[htbp]
\begin{center}
\begin{tabular}{l|l|rrr}
\toprule
 & \textbf{Model} &  \textbf{R@1}   & \textbf{R@5}  & \textbf{R@10}  \\ \midrule
1 & VinVL                     & 5.8  & 15.6 & 23.2  \\ 
2 & + $\textrm{template}_{base}$        & 7.1  & 19.3 & 27.2  \\ 
3 & + $\textrm{template}_{diverse}$     & 11.3 & 26.2 & 36.2 \\ \midrule
4 & VinVL w/ FT & 7.0  & 18.4 & 26.7   \\ 
5 & + $\textrm{template}_{base}$        &7.1 &19.5 &27.6  \\ 
6 & + $\textrm{template}_{diverse}$     &9.1 &23.0 &32.1 \\ 
\bottomrule
\end{tabular}
\end{center}
\caption{CLIP retrieval performance of template prompts on MSCOCO dataset. }
\label{tab:hd_results_all}
\end{table}

\begin{table*}[t]
\begin{center}
\begin{tabular}{l|l|ccc|c}
\toprule
\textbf{ImageID}                  & \textbf{Evaluated Sentences }                                                                   & \textbf{BLEU4} & \textbf{CIDEr} & \textbf{SPICE} & \textbf{CLIP sim} \\ \midrule
 &
  a group of people riding bikes down a city street \textcolor{blue}{, the sidewalk is busy}. &
 27.0 &
 97.0 &
 22.6 & 29.9\\ 
\multirow{-2}{*}{462565} &
  a group of people riding bikes down a city street \textcolor{blue}{with a busy sidewalk.} &
30.5 &
123.5 &
23.3 & 30.1\\ \midrule
                         & a young boy sitting on a bench in a park \textcolor{blue}{, the boy in yellow jacket.}         & 49.2  & 189.4 & 11.8 &33.7 \\ 
\multirow{-2}{*}{341921} & a young boy \textcolor{blue}{in yellow jacket} sitting on a bench in a park.                  & 57.8  & 233.8 & 11.8 & 32.6 \\  \midrule
                         & a bathroom with a toilet and a sink and a window \textcolor{blue}{, the bathroom with a basket.} & 32.8  & 83.9  & 13.3 & 29.9 \\
\multirow{-2}{*}{192440} & a bathroom with a toilet and a sink, a window \textcolor{blue}{and a basket. }                & 38.7  & 130.7 & 13.3 & 29.4 \\ \bottomrule
\end{tabular}
\end{center}
\caption{BLEU4, CIDEr, SPICE and CLIP similarity 
 scores on specific cases of MSCOCO test set that evaluate sentences with same semantics but different structures. The highlight words are detail-related. }
\label{tab:metrics}
\end{table*}

 \begin{figure*}[thbp]
    \centering
    \includegraphics[width=0.9\linewidth]{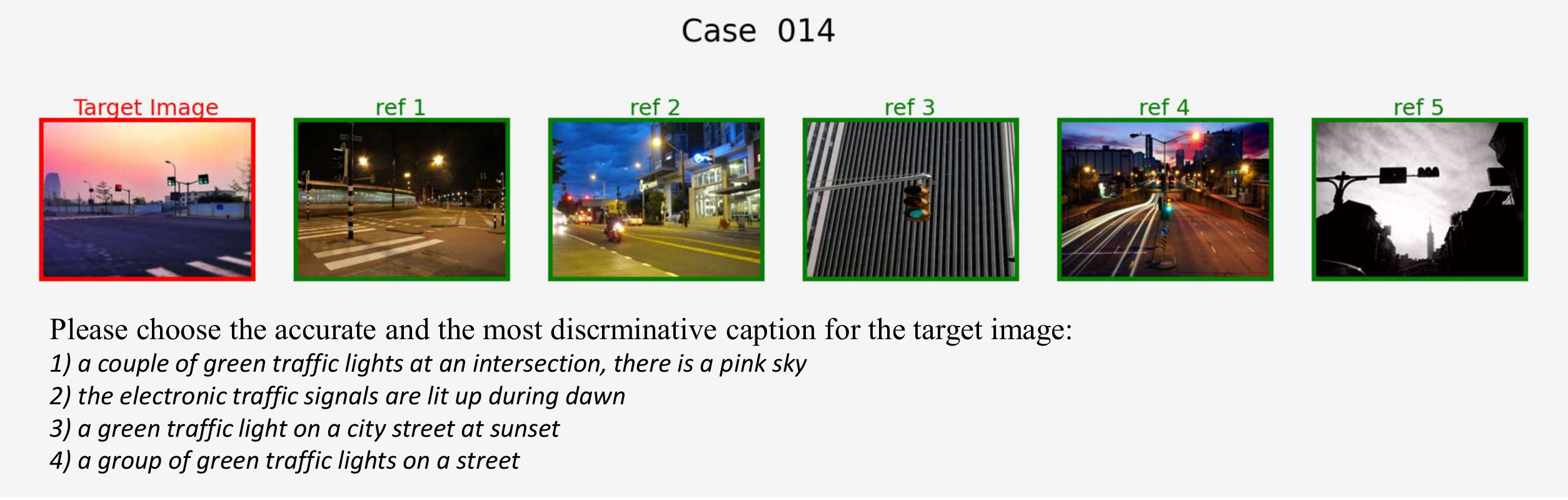}
    \caption{An example case to be evaluated in the user study.}
    \label{fig:user_study}
\end{figure*}

 \begin{figure*}[thb]
    \centering
    \includegraphics[width=\linewidth]{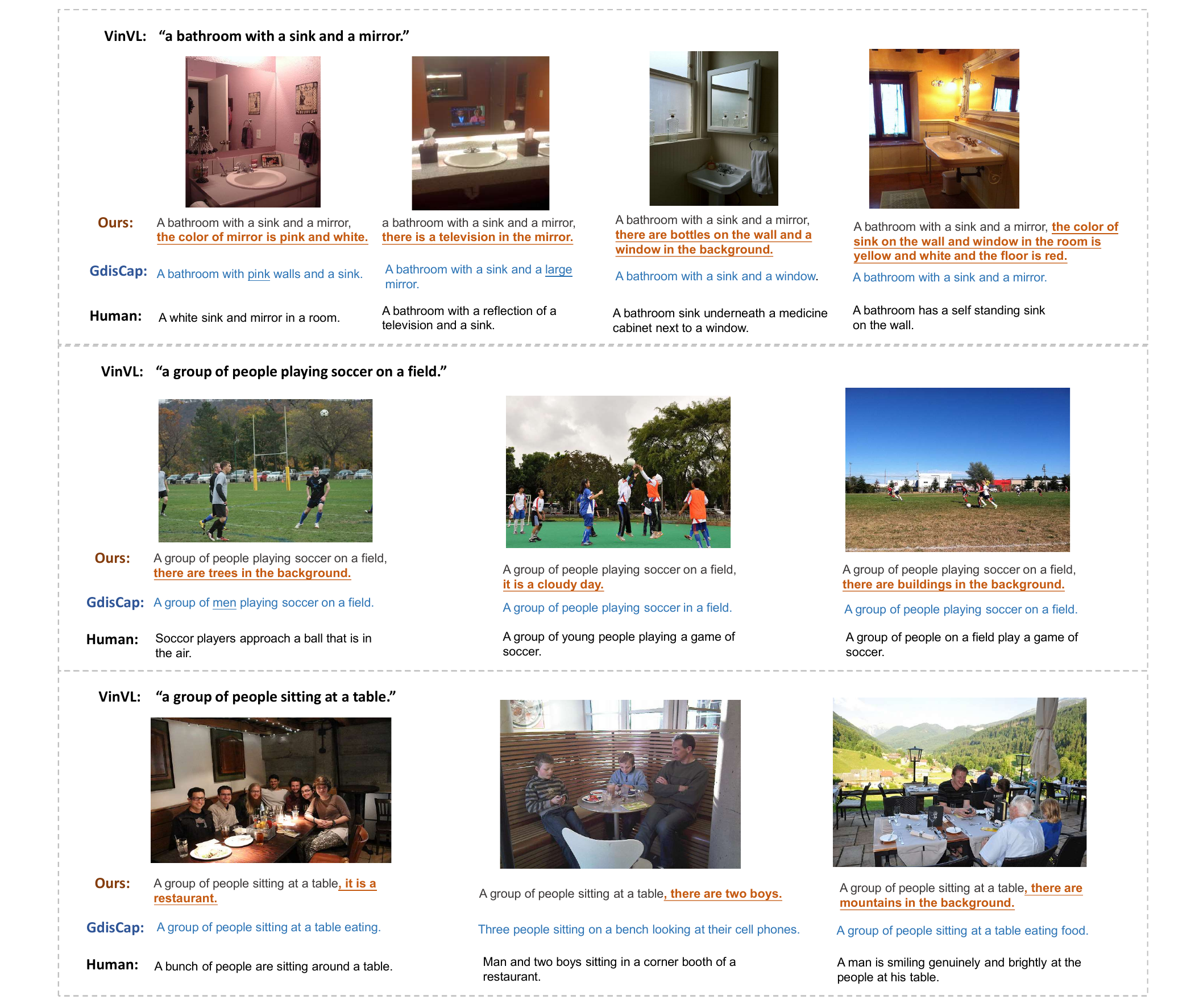}
    \caption{Additional visualization of captioning cases on MSCOCO test set. The newly generated details are underlined.}
    \label{fig:case_appendix}
\end{figure*}

\subsection{Analysis on N-gram Accuracy Metrics}
\label{sec:metric}
The conventional N-gram based metrics for image captioning, such as BLEU and CIDEr, are not suitable for evaluating the descriptions $\mathcal{Y}=\{\mathcal{G},\mathcal{D}\}$ generated by our framework.
It is due to the fact that our model generates paragraph-style of captions, which are structurally very different from the groundtruth captions. The N-gram based metrics like BLEU4 and CIDEr will penalize such structure difference although our captions are more descriptive, such phenomena was also reported in related works (Liu et al. 2019) as well. Table~\ref{tab:metrics} depicts some evaluated cases on MSCOCO test set. Descriptions with very same semantics but different sentence structure get very different metric scores, for example, the caption with sub-sentence structure leads to obvious degradation in CIDEr and BLEU4 scores compared with the caption with one-sentence structure, even though they have same semantics. Compared with CIDEr and BLEU, SPICE  is more robust to the structure variance of evaluated sentences because it first parses image captions into  scene graphs and then calculates the similarity between candidate and reference scene graphs. 
\textcolor{black}{Meanwhile, retrieval model CLIP can  eliminate the influence of sentence structure change and stably measure the image-text semantics similarity. Therefore, CLIP-based accuracy metrics, i.e. CLIPScore and RefCLIP-Score, which directly measure the whole semantic consistence of sentences and images, can also mitigate the structural impacts.}
 Based on the above observation, we utilize SPICE, CLIPScore and  RefCLIP-Score as \textit{accuracy} metrics.  For more reliable evaluation of generated sentences, we also conduct user study in Section 5.3 (Figure 5), and the human evaluation results  are consistent with the tendency of automatic metrics shown in Section 4.3 (Table 1).

\subsection{User Study Interface}
\label{sec:user_study}
Figure~\ref{fig:user_study} illustrates an example case in the user study. For each target image, users need to select the accurate and most descriptive caption among four candidates generated by different models.
 In each case, we also provide five additional images that have similar visual scenarios to the target image as a reference. Besides, each participant is asked to focus on the semantic accuracy and descriptiveness regardless of the sentence structure.
 
\subsection{\textcolor{black}{Additional Visualized Cases}}
We present more generated descriptions of similar image sets on the MSCOCO test set in Figure~\ref{fig:case_appendix}.